\pdfoutput=1
\documentclass[letterpaper, 10 pt, conference]{ieeeconf}
\usepackage{pdfpages}
\usepackage{amsmath}
\usepackage{amsfonts}
\usepackage{multirow}
\usepackage{here}
\usepackage{bm}
\usepackage{enumerate}
\usepackage{tabularx}
\usepackage{xcolor}
\usepackage{url}
\usepackage{color}
\usepackage{longtable}
\usepackage[subrefformat=parens]{subcaption}
\usepackage{algorithm}
\usepackage{algorithmic}
\usepackage{threeparttable}

\def\rnum#1{\expandafter{\romannumeral #1}}
\def\Rnum#1{\uppercase\expandafter{\romannumeral #1}}
\newcommand{\argmax}{\mathop{\rm arg~max}\limits}
\newcommand{\argmin}{\mathop{\rm arg~min}\limits}

\IEEEoverridecommandlockouts
\title{\LARGE \bf
Video Motion Capture from the Part Confidence Maps \\ of Multi-Camera Images by Spatiotemporal Filtering \\ Using the Human Skeletal Model
}

\author{Takuya Ohashi, Yosuke Ikegami, Kazuki Yamamoto, Wataru Takano and Yoshihiko Nakamura
\thanks{This research was supported by Japan Society for the Promotion of Science under Category(A) of Grants-in-Aid for Scientific Research(17H00766), "Sports Simulation Science for Intervention in Acquisition and Learning of Motion"(PI: Y.Nakamura)}
\thanks{T. Ohashi, Y. Ikegami, K. Yamamoto and Y. Nakamura are with the Department of Mechano-Informatics, The University of Tokyo, 7-3-1 Hongo, Bunkyo-ku, 113-8656, Tokyo, Japan. ohashi, ikegami, yamamoto, nakamura@ynl.t.u-tokyo.ac.jp}
\thanks{W. Takano is with the Center for Mathematical Modeling and Data Science, Osaka University, 1-3 Machikaneyamacho, Toyonaka-shi, Osaka, Japan. takano@sigmath.es.osaka-u.ac.jp}
}

\begin{document}

\maketitle
\thispagestyle{empty}
\pagestyle{empty}

\begin{abstract}
This paper discusses video motion capture, namely, 3D reconstruction of human motion from multi-camera images.
After the Part Confidence Maps are computed from each camera image, the proposed spatiotemporal filter is applied to deliver the human motion data with accuracy and smoothness for human motion analysis.
The spatiotemporal filter uses the human skeleton and mixes temporal smoothing in two-time inverse kinematics computations. The experimental results show that the mean per joint position error was 26.1mm for regular motions and 38.8mm for inverted motions.

\end{abstract}

\section{INTRODUCTION}

Motion capture technology is widely used for many applications, in particular with human motion, such as animation, video game production, sports training, biomechanical analysis, rehabilitation, medical diagnosis, and research of human motion modeling and recognition \cite{sdims2}\cite{Murai1}\cite{Takano:2014}.

The technology has three typical methods for measurements, namely, passive optical, active optical, and gyro-accelerometer measurements. The spatiotemporal accuracy is approximately 1mm and 1kHz for the typical human scale environments in the case of the high-end passive optical system \cite{MotionA}\cite{Vicon}, and getting even higher by the advancement of imaging device technology. The passive optical method imposes the least constraint to a human subject, but still needs approximately 40 spherical reflective markers attached on the body. Other methods also have similar or even stronger constraints.

To solve this problem, image-based motion capture which imposes no constraint to the human subjects is demanded.
The studies on RGB-D vision show promising results for human motion recognition using Kinect \cite{Shotton1}\cite{Tong1} and the other systems with laser projection or scanning depth sensors \cite{Spinello}\cite{Dewan}. Recent studies on single RGB vision \cite{Mehta1}\cite{VNect:2017}\cite{Kanazawa1}\cite{Sun1} show remarkable results where, taking an advantage of the large RGB image data set and the power of deep neural networks, the learning based methods for human motion recognition have been developed.

Although image-based human motion recognition is considered a 2D problem and rather tolerant in spatial accuracy and temporal discontinuity, motion capture is a 3D reconstruction problem and requires spatiotemporal accuracy and smoothness for velocity and acceleration analysis.

In this paper, we report about our study on video motion capture, which means image-based 3D human motion reconstruction with spatiotemporal accuracy and smoothness closer to passive optical motion capture.
We use synchronized multiple cameras to record video images of a human subject from different directions. The likelihood of joint positions are computed as the Part Confidence Maps (PCM) \cite{Wei:2016}\cite{Cao:2017} in each video image and provided by the colors of heatmap with the labels of individual joints by OpenPose \cite{OpenPose}. Our initial study showed that the simple least square solution of the 3D position of each joint from the multi-camera images neither resulted in smooth 3D joint movements nor maintained link length between adjacent joints.

We propose to use the human skeletal model for reconstructing 3D human motion by spatiotemporal filtering of joint movements. We also discuss to enhance function of the PCM for rather difficult human postures. The proposed video motion capture was implemented and tested through various experiments. The results of evaluation are shown based on the reconstructions made by two methods, namely, the video motion capture and the passive optical motion capture.

\section{Video Motion Capture from Multi-Camera Images}
\subsection{Flow of the Video Motion Capture}
\begin{figure}[H]
\begin{center}
\includegraphics[width=0.93\hsize]{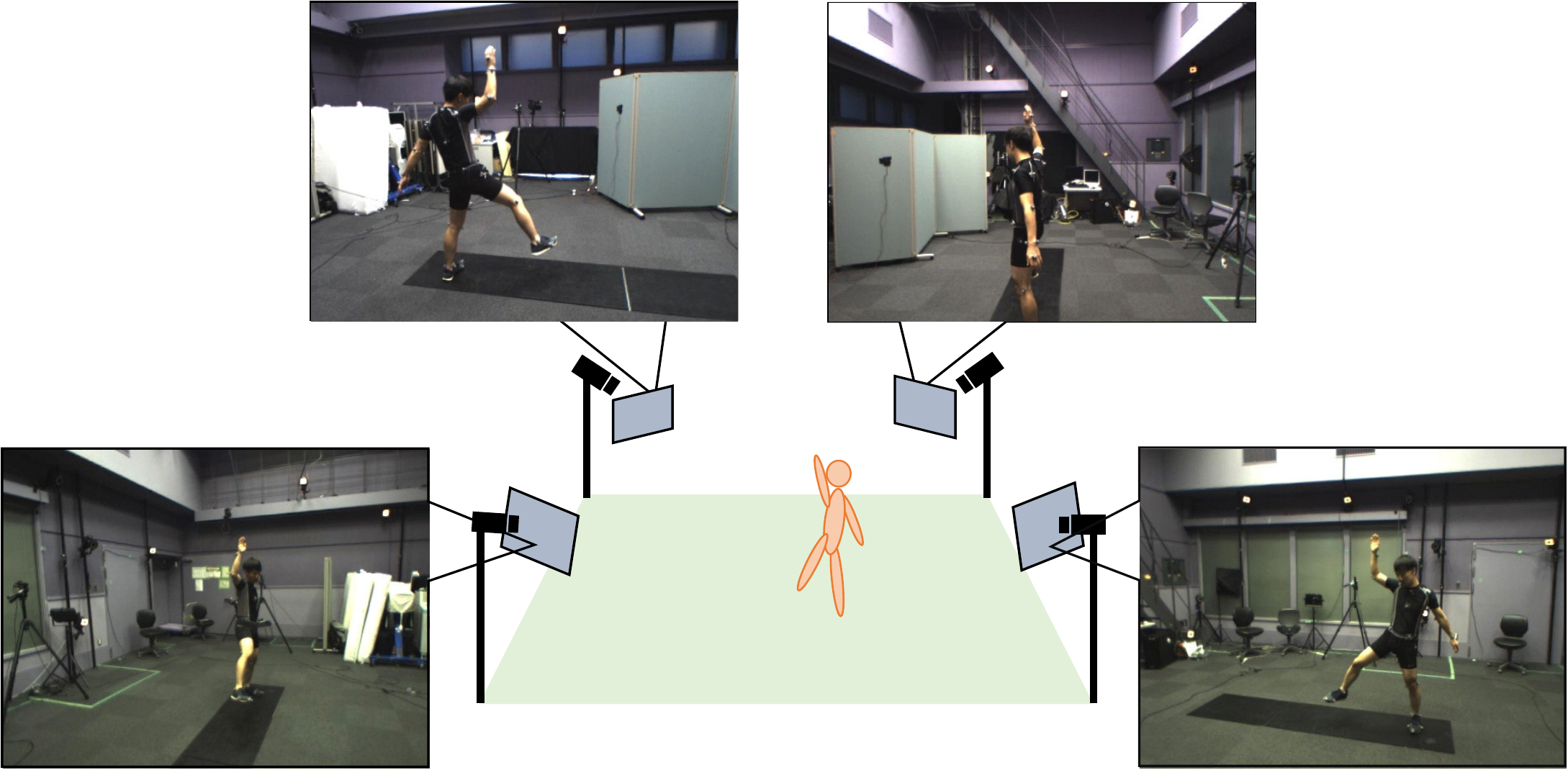}
\caption{Measurement Environment}
\label{fig:camera_pos}
\end{center}
\end{figure}

\begin{figure}[H]
\begin{center}
\includegraphics[width=0.8\hsize]{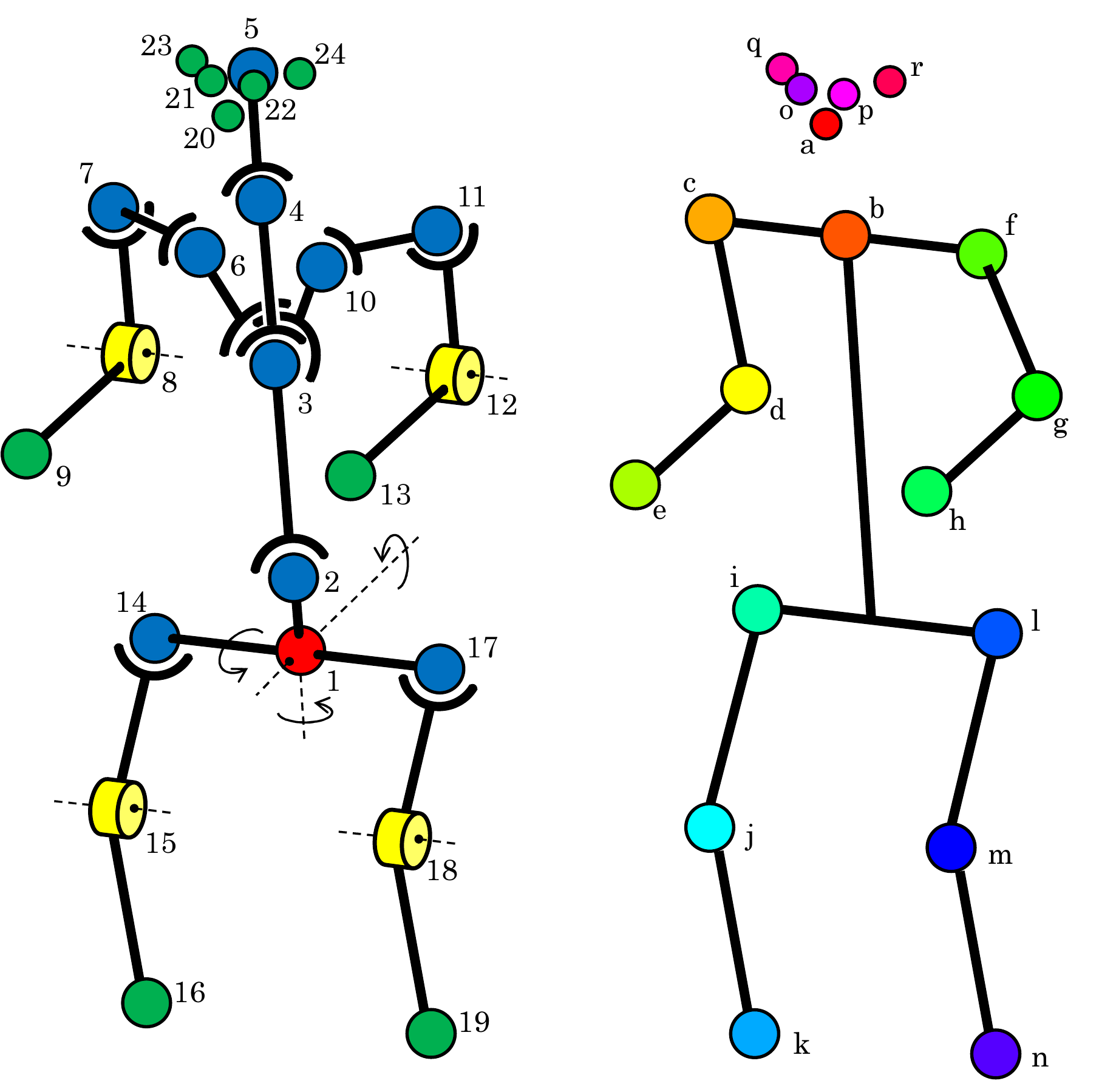}
\hspace{6cm} [a] 40DOF Skeletal Model \hspace{0.4cm} [b] Virtual Markers
\caption{Correspondence of Human Skeletal Model and Virtual Markers. In [a], joints color indicates as 6DOF (red), 3DOF (blue) and 1DOF (yellow)}
\label{fig:human_model}
\end{center}
\end{figure}

Proposed video motion capture can be described as the iterative update of human skeletal model's joint angles based on multi-camera images which surround a human subject like Fig.\ref{fig:camera_pos}.
Note that in this paper the skeletal model means a virtual open tree-structure kinematic chain with 40 degrees of freedom (DOF) as shown in Fig.\ref{fig:human_model} [a].
To fulfil this update, the weighted markers which are virtually generated in the 3D space are computed as shown in Fig.\ref{fig:human_model} [b] based on the subject's probable joint positions.
By minimizing the errors of virtual markers' positions and  skeletal model's corresponding joint positions, the skeletal model's joint positions at the time are obtained.
To solve this minimisation problem, inverse kinematics (IK) computation is adapted, and by repeating above process, the video motion capture is realized.

\begin{figure}[t]
\begin{center}
\includegraphics[width = 0.9\hsize]{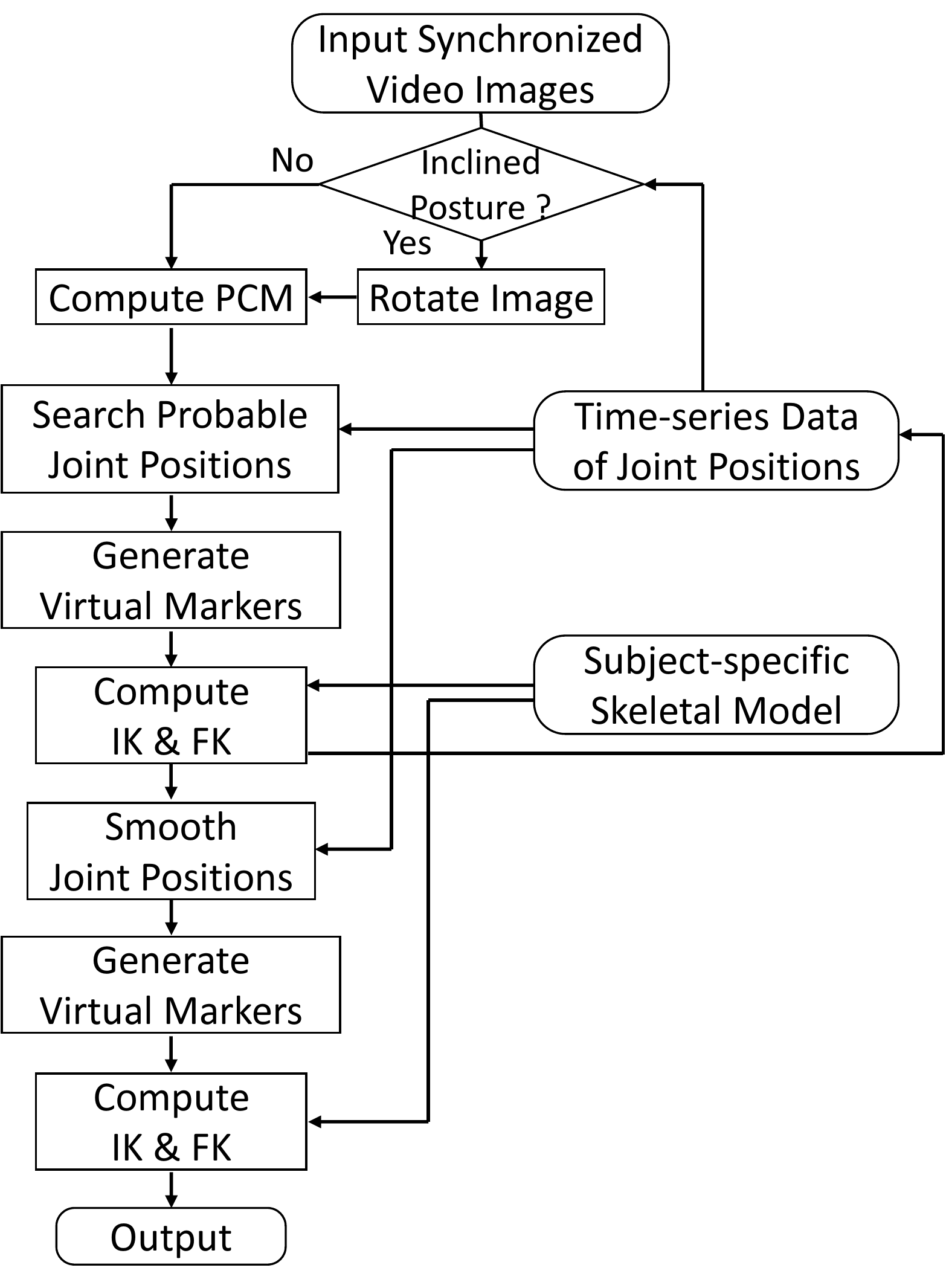}
\caption{Flow Chart of the Video Motion Capture}
\label{fig:flow}
\end{center}
\end{figure}

The flow chart of the method is shown in Fig.\ref{fig:flow}.
Note that the camera calibration must be done before the computation.
The camera parameters such as the focal length, the center of image, and the lens distortion enable restoration of distorted images and provide the transformation matrices $\mathcal P \it _i$ that maps an arbitrary 3D point to the corresponding pixel location on the $i$th camera image plane \cite{Zhang:2000}\cite{opencv}.
Other main blocks in the chart are explained in the following subsections.

\subsection{Computation of the Part Confidence Maps}

The OpenPose \cite{OpenPose} provides the Part Confidence Maps (PCM) \cite{Wei:2016} and the Part Affinity Fields (PAF) \cite{Cao:2017} of multi-person's keypoints on a 2D image in real time using the trained Convolutional Neural Networks (CNN).
The keypoints are 18 consisting of joint positions like shoulders, elbows, wrists, hips, knees, ankles and a neck, and the other feature points like eyes, ears, and a nose.
Though the recent version of OpenPose includes hands, feet, and face, we used only the above 18 keypoints.

\begin{figure}[H]
\begin{center}
\includegraphics[width=\hsize]{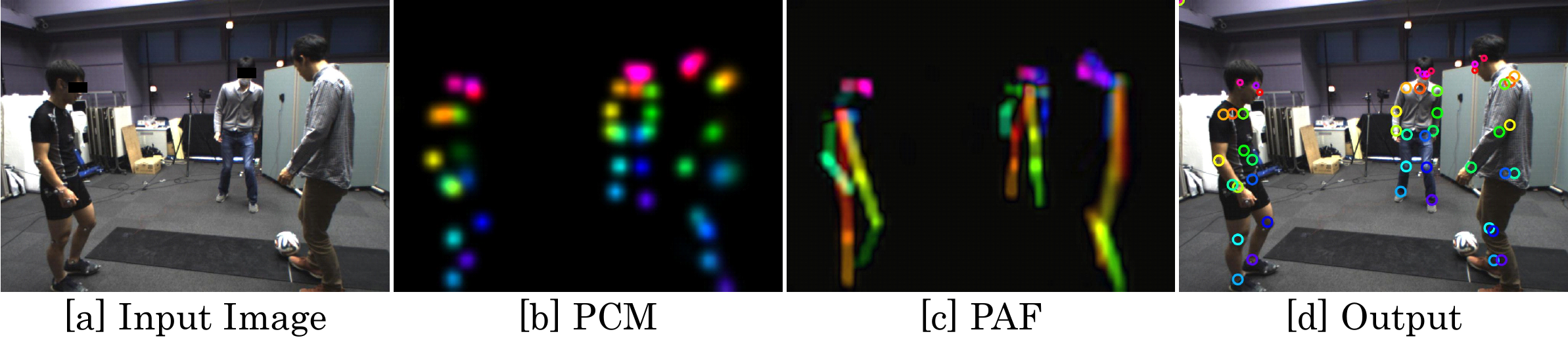}
\caption{Multi-Person Keypoints Detection Using OpenPose}
\label{fig:openpose_output}
\end{center}
\end{figure}

The PCM shows by its temperature the distribution of likelihood of the keypoint, namely, in each pixel location, the probability of the keypoint is expressed as the continuous value from 0 to 1.
The centroid of the PCM may be considered as the position of the corresponding keypoint on the 2D image.
Note that we use the PCM not its centroids for the 3D human motion reconstruction.
The PAF shows the connectivity of the keypoints of a single person. In this paper, we do not use the PAF though it is a powerful means to identify individuals of multi-person on the image. An example of input image [a], PCM [b], PAF [c], and the centroids of PCM superimposed on the image [d] are shown in the Fig.\ref{fig:openpose_output}

\subsection{Initialization of Skeletal Model and Joint Positions}

In order to compute IK, the skeletal model's adjacent joints must be connected by a constant-length link and the link length must be calculated according to the human subject.
Furthermore, since proposed motion capture is based on iterative computation, it is reasonable to calculate the skeletal model's initial joint positions before the computation.

\begin{figure}[H]
\begin{center}
\includegraphics[width = 0.9\hsize]{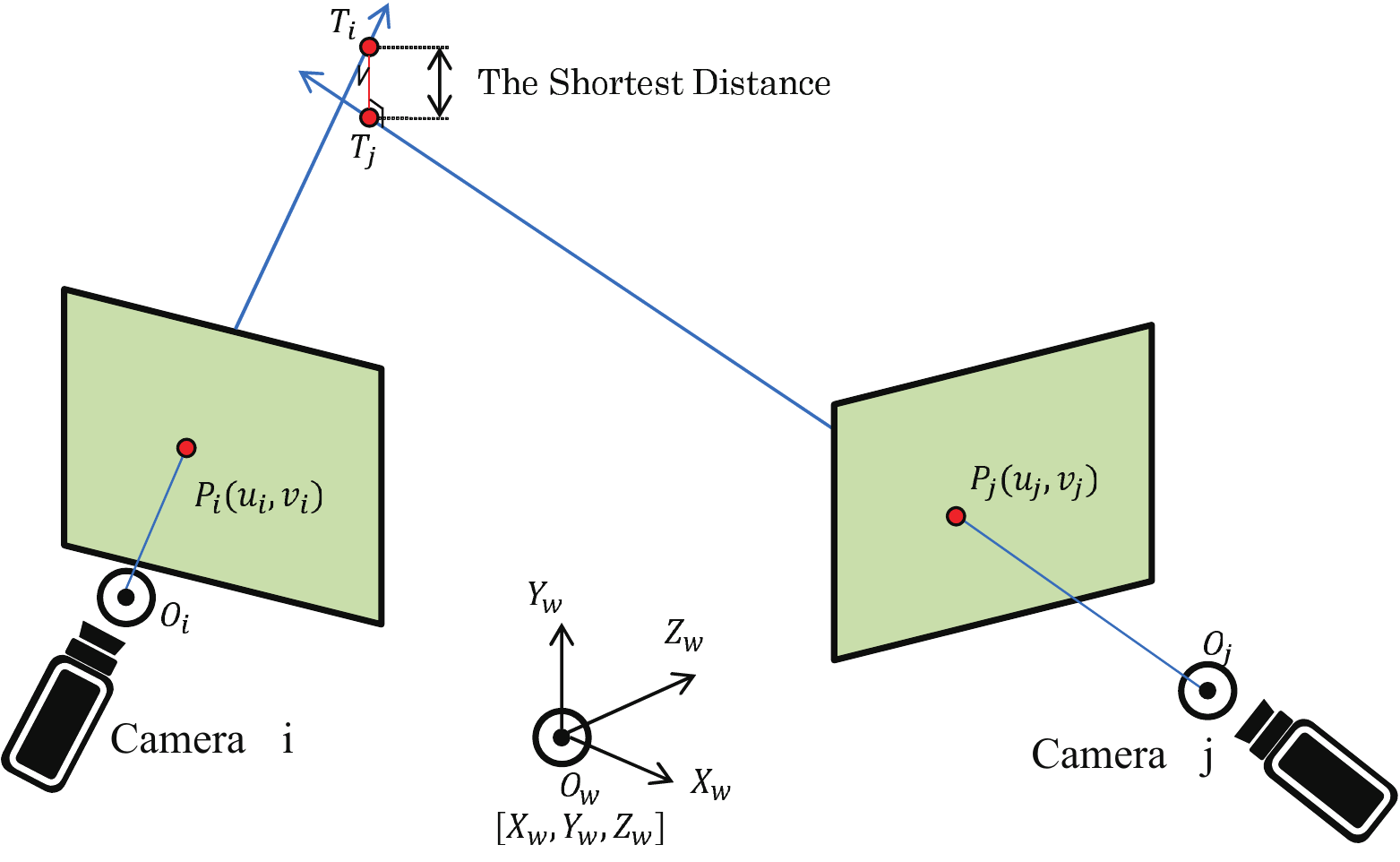}
\caption{3D Position Calculation from Pixel Location}
\label{fig:shortest}
\end{center}
\end{figure}

The link lengths of the skeletal model are identified for the subject by using the centroids of PCM on the images from multiple cameras as schematically shown in Fig.\ref{fig:shortest}.
Initialization is confirmed when the centroids of every keypoint from the cameras show a 3D agreement.
And simultaneously, skeletal model's initial joint positions are calculated.

\subsection{Searching Probable Joint Positions}
Let $^{t}J^n$ and $^{t}Q^n$ represent the 3D joint position and the joint angle of $n$th joint at $t$ time-frame.
We discretize the 3D space with a unit distance $s$, and consider the lattice of $(2k+1) \times (2k+1) \times (2k+1)$ with $^{t}J^n$ at the center. We then search for $^{t+1}J^n$ on the lattice. The search space $^{t}\mathbb{J}^n$ is represented by

\begin{equation}
\begin{split}
    ^{t}\mathbb{J}^n :=
     \begin{Bmatrix}
     ^{t}J^n + s \left.
    \begin{bmatrix}
    a\\
    b\\
    c\\
    \end{bmatrix} \hspace{0.2cm} \right| \hspace{0.2cm}
    &-k \leq a,b,c \leq k
    \end{Bmatrix} \\
    k: \rm constant \hspace{0.06cm} positive \hspace{0.06cm} integers \\\it a,b,c: \rm integers
\label{eq:cube}
\end{split}
\end{equation}

A point on the lattice in $^{t}\mathbb{J}^n$ is described as $^{t}J_{a,b,c}^n$.
It is considered that the point with the highest score from summation of multi-camera PCM values is the most probable point of the joint existence.
Let $n_{c}$ be the number of cameras and the function $^{t+1}\mathcal S \it _i^n$ be the PCM value that of a corresponding pixel positionof $n$th joint on $i$th camera image at $t+1$.
Then, the predicted joint position at $t+1$ can be calculated as follows:

\begin{equation}
    ^{t+1}J^n_{pred} = \argmax_{-k \leq a,b,c \leq k} \sum_{i=1}^{n_{c}} {^{t+1}\mathcal S \it _i^n(\mathcal P \it _i(^{t}J_{a,b,c}^n))}
\label{eq:cube_search}
\end{equation}

At this predicted joint position, virtual marker is generated.

\subsection{Computing Inverse Kinematics}

Since the joint position $^{t+1}J^n$ is a function of joint angle $^{t+1} \bf Q$ expressed in forward kinematics (FK), $^{t+1} \bf Q$ can be obtained by solving following optimization problem \cite{Ayusawa1}:

\begin{equation}
    ^{t+1} \bf Q \it = \argmin \sum_n^{\rm n_{j} \it} \cfrac{\rm 1 \it}{\rm 2 \it} {^{t+1}W^n}  ||^{t+1}J^n _{pred} - ^{t+1}J^n||^{\rm 2 \it}
\end{equation}
\noindent
Note that $n_{j}$ is the number of joints and the constant weight $^{t+1}W^n$ is obtained from the PCM score at the predicted joint position as follows.
\begin{equation}
    ^{t+1}W^n = \sum_{i=1}^{n_{c}} {^{t+1}\mathcal S \it _i^n(\mathcal P \it _i(^{t+1} J^n _{pred}))}
\label{eq:PCM_SCORE}
\end{equation}
\noindent

\subsection{Smoothing Joint Positions with Maintaining Link Length}
Human movement is smooth and differentiable by time, but acquisition of PCM and IK computation does not use time-series data, so the output joint data is not to be smoothed.
To solve this problem, joint position is smoothed by introducing IIR low-pass filter.
However, the time-invariant parameters of skeletal model, link length, is changed every time-frame by the smoothing.

In proposed method, the position of the joint after the adaptation of the low-pass filter is set as the virtual marker of each joint, and the optimization by IK computation is performed again.
This makes it possible to make use of the smoothness of the temporal change of the joint position under the skeletal condition that each link length is invariant.

\section{Experiments}

\subsection{Comparison with Optical Motion Capture}

\begin{figure}[H]
\begin{center}
 \begin{minipage}[b]{0.48\linewidth}
  \centering
  \includegraphics[width=\hsize]
  {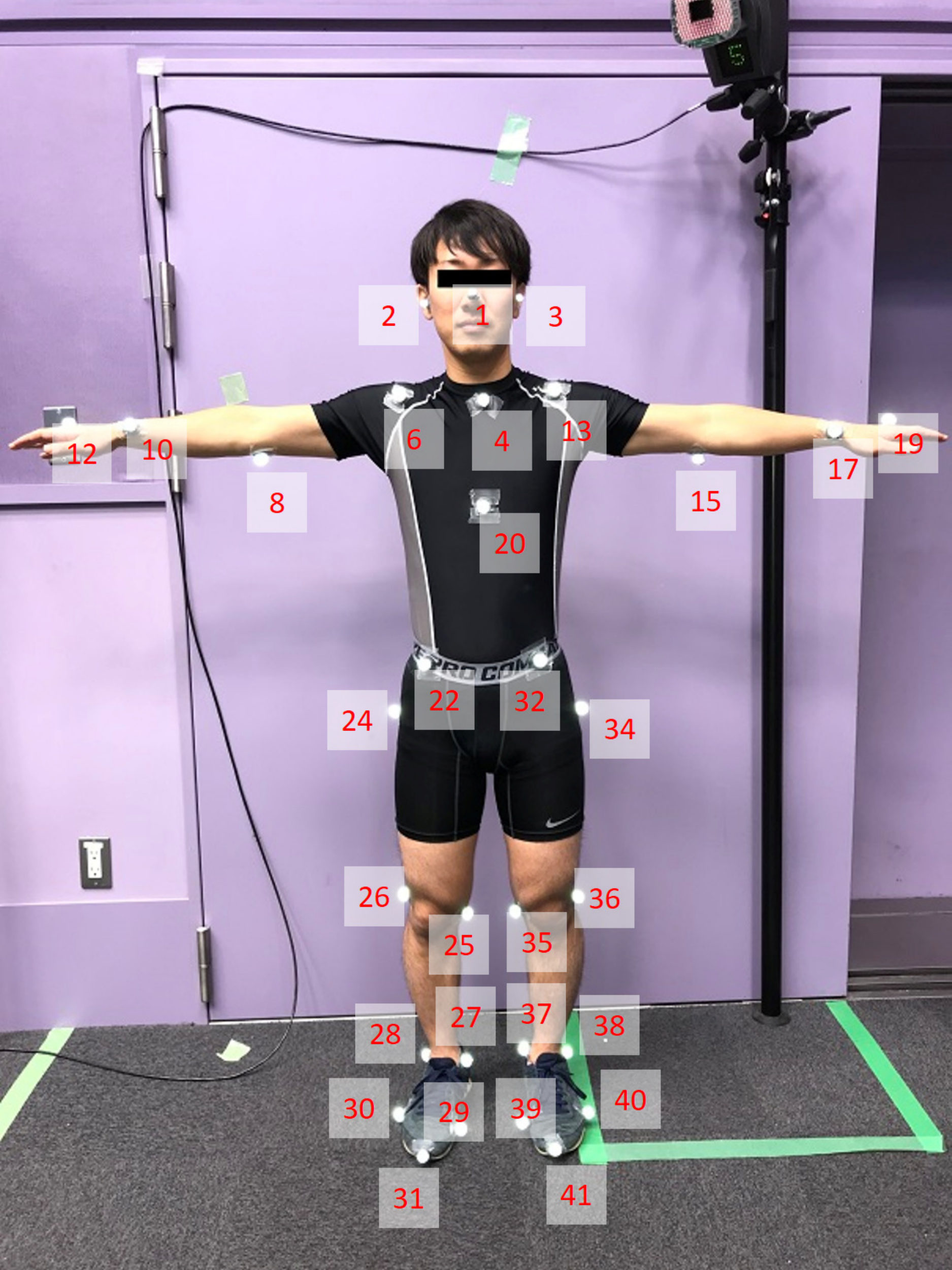}
 \end{minipage}
 \begin{minipage}[b]{0.48\linewidth}
  \centering
  \includegraphics[width=\hsize]
  {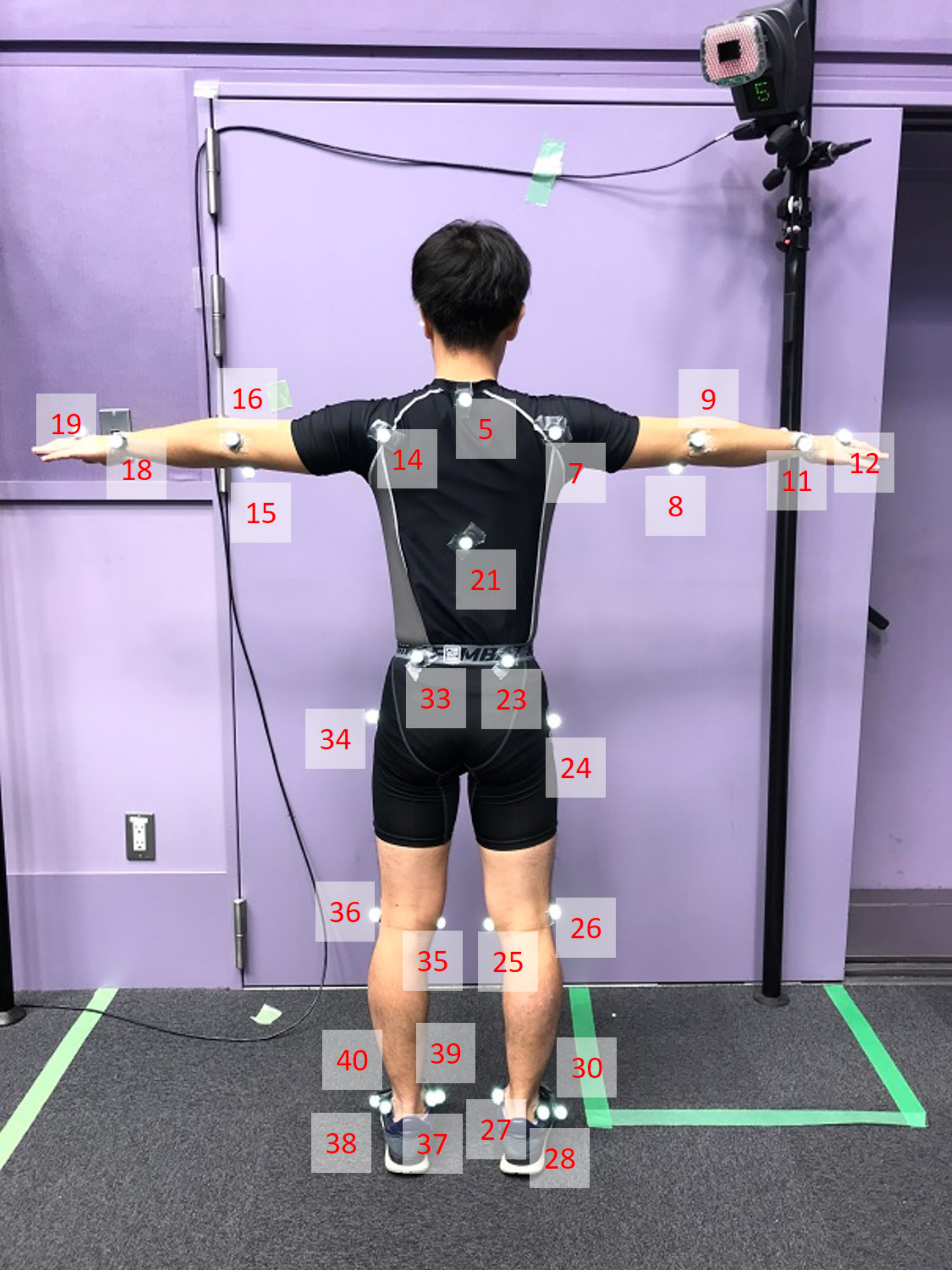}
 \end{minipage}
\caption {Positions of the Attached Markers}
\label{fig:optical_markers}
\end{center}
\end{figure}

Human subject's motion was synchronously recorded by 4 RGB cameras (FLIR Corporation, Grasshopper3) at 60Hz with a 1024$\times$768 pixel resolution and measured by the optical motion capture system with 17 infra-red cameras (Motion Analysis Corporation, Eagle and Raptor-4) at 200Hz with 41 markers.
Fig.\ref{fig:optical_markers} shows the positions of the attached markers.
Ground truth of joint positions was calculated by these measured markers.

In this experiment, 9 motions were measured; Walk, Dance, Spin Jump, Football Juggling, Golf, Matrix, Handstand, Cartwheel, and Break Dance.

\begin{figure}[H]
 \begin{minipage}[b]{0.99\linewidth}
  \centering
  \includegraphics[width=\hsize]
  {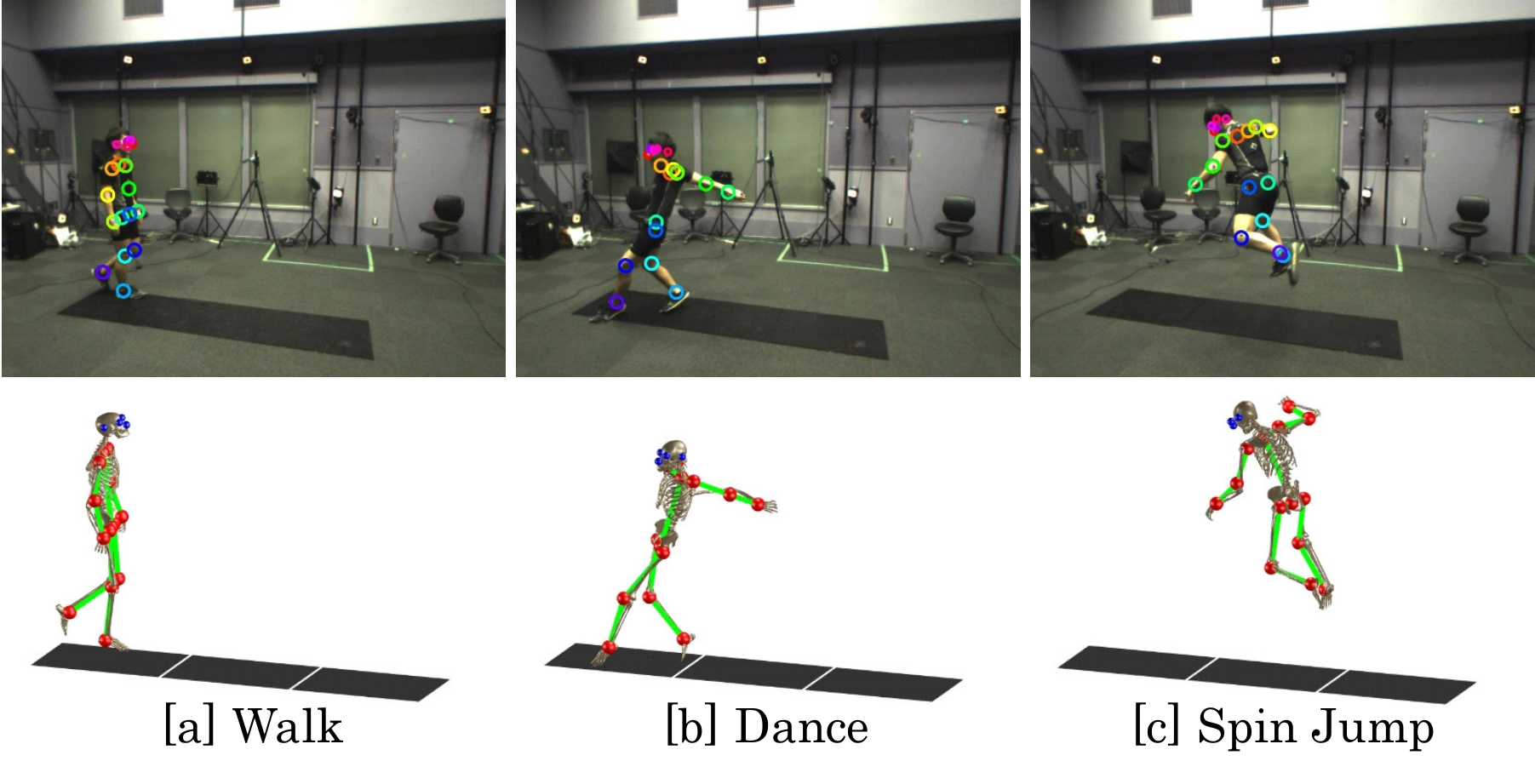}
 \end{minipage}
 \begin{minipage}[b]{0.99\linewidth}
  \centering
  \includegraphics[width=\hsize]
  {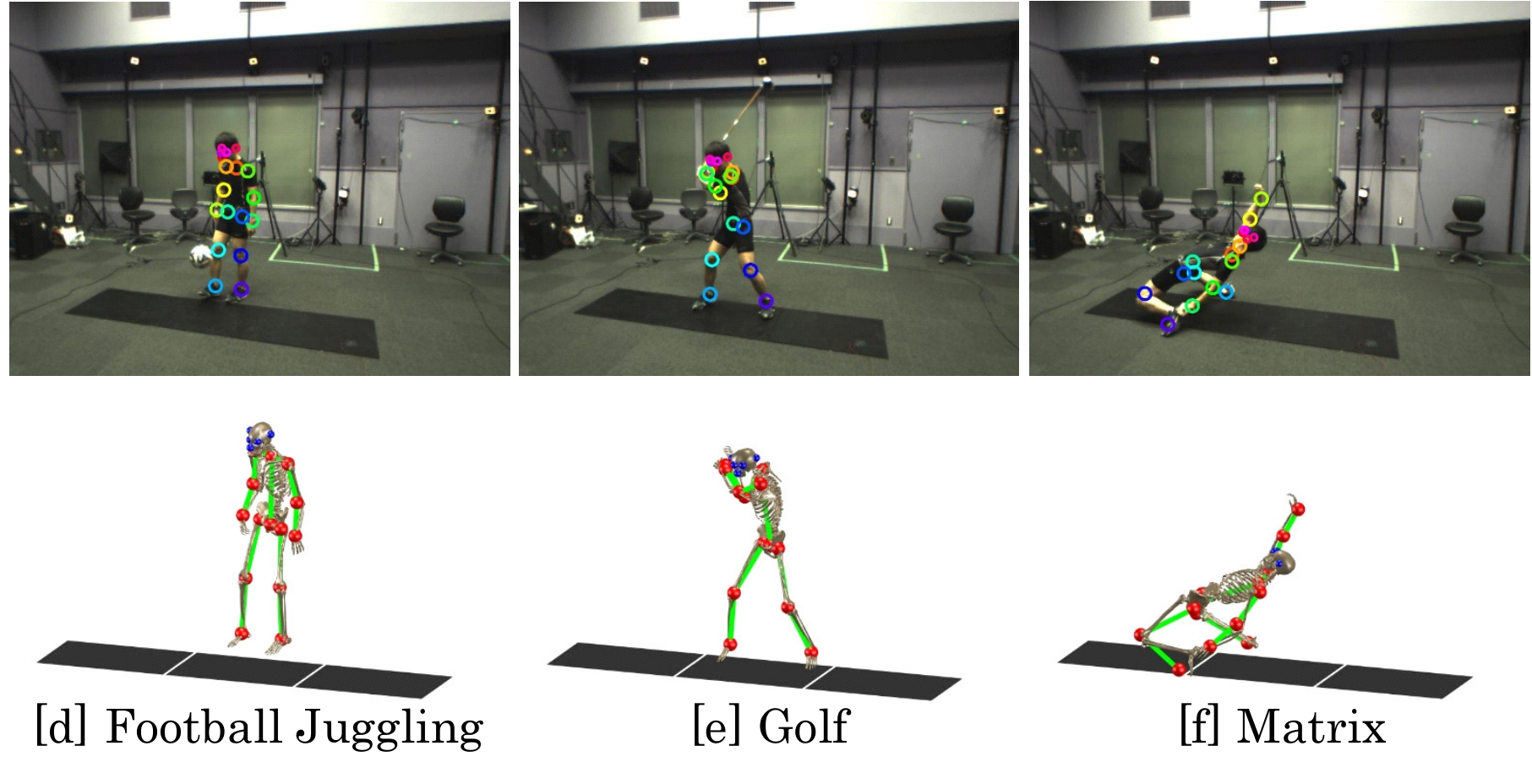}
 \end{minipage}
 \begin{minipage}[b]{0.99\linewidth}
  \centering
  \includegraphics[width=\hsize]
  {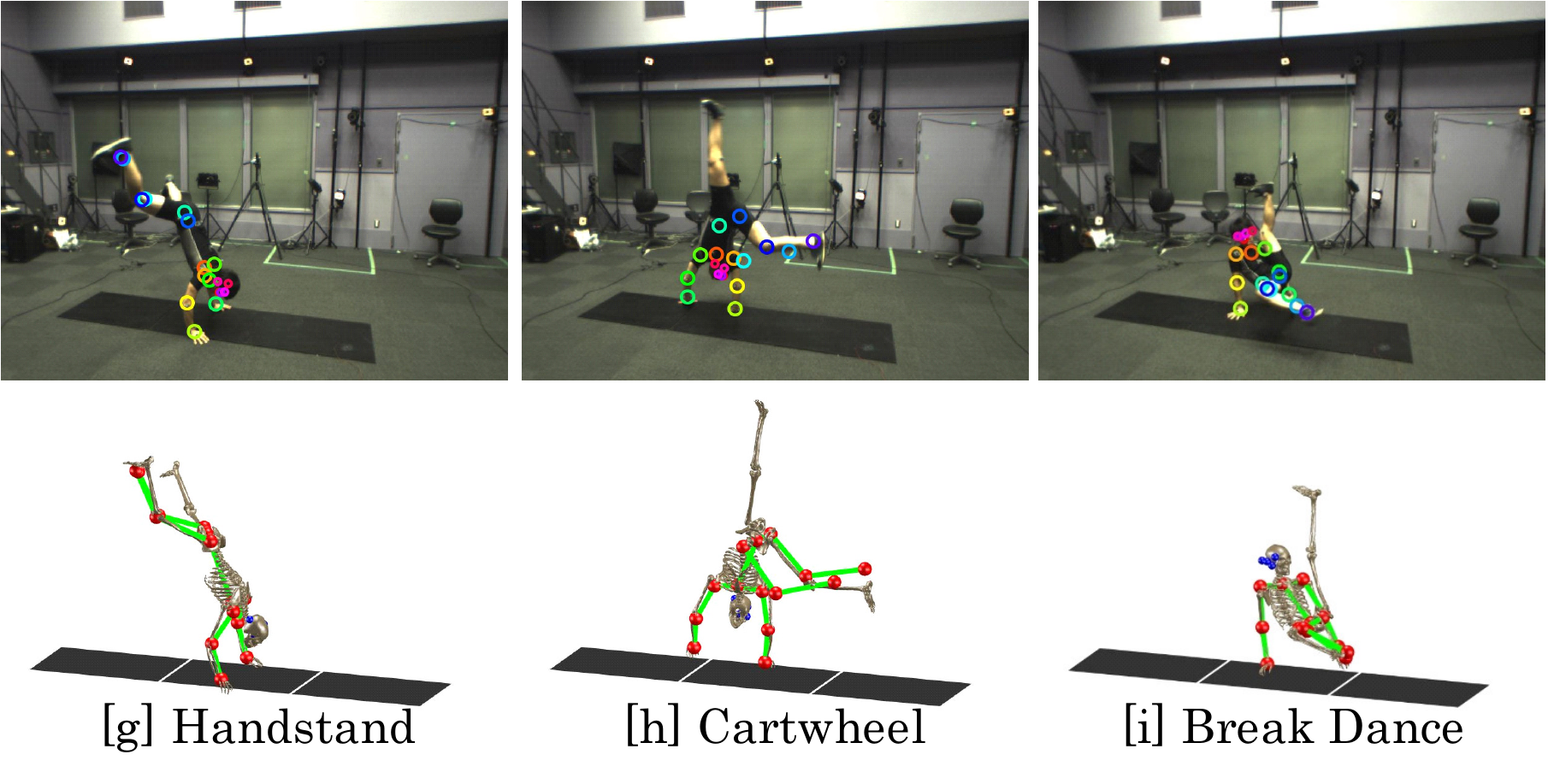}
 \end{minipage}
\caption {Results of the Video Motion Capture}
\label{fig:output}
\end{figure}

Fig.\ref{fig:output} shows representative frame of each motion. Top images represent input images. Estimated joint positions from the video motion capture are drawn as circles.
Bottom images represent 3D pose. Joint positions from the video motion capture are drawn as stick model and joint positions from optical motion capture are drawn as bone model.

Table.\ref{tab:mpjpe1} shows the results of the mean per joint position error (MPJPE) of each motion and percentage of correct keypoints in 3D (3D-PCK) when all the joints are in the images.
UpperBody consists of neck, shoulders, elbows, and wrists. LowerBody consists of hips, knees, and ankles. Head consists of nose and ears.

\begin{table*}
\centering
\caption{Accuracy of Video Motion Capture}
\label{tab:mpjpe1}
\begin{tabular}{|l|l||c|c|c|c|c|c|c|c|c||c|}
\hline
\multicolumn{2}{|l||}{} & \multicolumn{1}{l|}{Walk} & \multicolumn{1}{l|}{Dance} & \multicolumn{1}{l|}{\begin{tabular}[c]{@{}l@{}}Spin\\ Jump\end{tabular}} & \multicolumn{1}{l|}{\begin{tabular}[c]{@{}l@{}}Football\\ Juggling\end{tabular}} & \multicolumn{1}{l|}{Golf} & \multicolumn{1}{l|}{Matrix} & \multicolumn{1}{l|}{Handstand} & \multicolumn{1}{l|}{Cartwheel} & \multicolumn{1}{l||}{\begin{tabular}[c]{@{}l@{}}Break\\ Dance\end{tabular}} & \multicolumn{1}{l|}{Total} \\ \hline \hline
\multirow{4}{*}{\begin{tabular}[c]{@{}l@{}}MPJPE\\ {[}mm{]}\end{tabular}} & Head & 19.9 & 25.4 & 29.8 & 21.3 & 22.9 & 32.4 & 38.5 & 37.2 & 33.1 & 26.9 \\ \cline{2-12}
 & UpperBody & 29.7 & 29.3 & 31.4 & 20.3 & 32.4 & 28.5 & 29.1 & 34.3 & 32.4 & 29.6 \\ \cline{2-12}
 & LowerBody & 26.0 & 23.0 & 23.7 & 25.0 & 18.4 & 30.7 &  96.0 & 81.7 &  57.4 & 33.3 \\ \cline{2-12}
 & Total & 26.5 & 26.2 & 28.2 & 22.3 & 25.3 & 30.1 &  56.0 & 52.6 & 41.9 & 30.5 \\ \hline \hline
 \multirow{4}{*}{\begin{tabular}[c]{@{}l@{}}3D-PCK\\ {[}\%{]}\\ @ 50mm\end{tabular}} & Head & 100 & 99.8 & 95.4 & 100 & 99.8 & 88.7 & 79.7 & 85.6 & 88.6 & 95.9 \\ \cline{2-12}
 & UpperBody & 99.3 & 93.0 & 91.8 & 99.9 & 87.0 & 92.5 & 89.4 & 78.6 & 90.1 & 92.1 \\ \cline{2-12}
 & LowerBody & 98.8 & 97.3 & 94.5 & 99.0 & 100 & 92.2 & 72.1 & 87.0 & 82.4 & 94.2 \\ \cline{2-12}
 & Total & 99.3 & 95.9 & 93.5 & 99.6 & 94.3 & 91.7 & 81.1 & 83.1 & 86.9 & 93.6 \\ \hline \hline
\multirow{4}{*}{\begin{tabular}[c]{@{}l@{}}3D-PCK\\ {[}\%{]}\\ @ 100mm\end{tabular}} & Head & 100 & 100 & 100 & 100 & 100 & 100 & 98.5 & 93.7 & 99.4 & 99.6 \\ \cline{2-12}
 & UpperBody & 100 & 99.8 & 99.9 & 100 & 98.6 & 100 & 100 & 98.0 & 98.8 & 99.5 \\ \cline{2-12}
 & LowerBody & 100 & 100 & 99.5 & 100 & 100 & 100 & 81.4 & 90.1 & 92.4 & 97.8 \\ \cline{2-12}
 & Total & 100 & 99.9 & 99.8 & 100 & 99.4 & 100 & 92.7 & 94.2 & 96.5 & 98.9 \\ \hline \hline
\multirow{4}{*}{\begin{tabular}[c]{@{}l@{}}3D-PCK\\ {[}\%{]}\\ @ 150mm\end{tabular}} & Head & 100 & 100 & 100 & 100 & 100 & 100 & 99.9 & 96.3 & 100 & 99.8 \\ \cline{2-12}
 & UpperBody & 100 & 99.9 & 99.9 & 100 & 99.4 & 100 & 100 & 100 & 99.3 & 99.8 \\ \cline{2-12}
 & LowerBody & 100 & 100 & 99.8 & 100 & 100 & 100 & 86.2 & 92.0 & 95.3 & 98.4 \\ \cline{2-12}
 & Total & 100 & 99.9 & 99.9 & 100 & 99.8 & 100 & 94.8 & 96.3 & 97.9 & 99.3 \\ \hline
\end{tabular}
\end{table*}

\begin{figure}[H]
\begin{center}
\includegraphics[width=\hsize,bb=0 0 660 328]{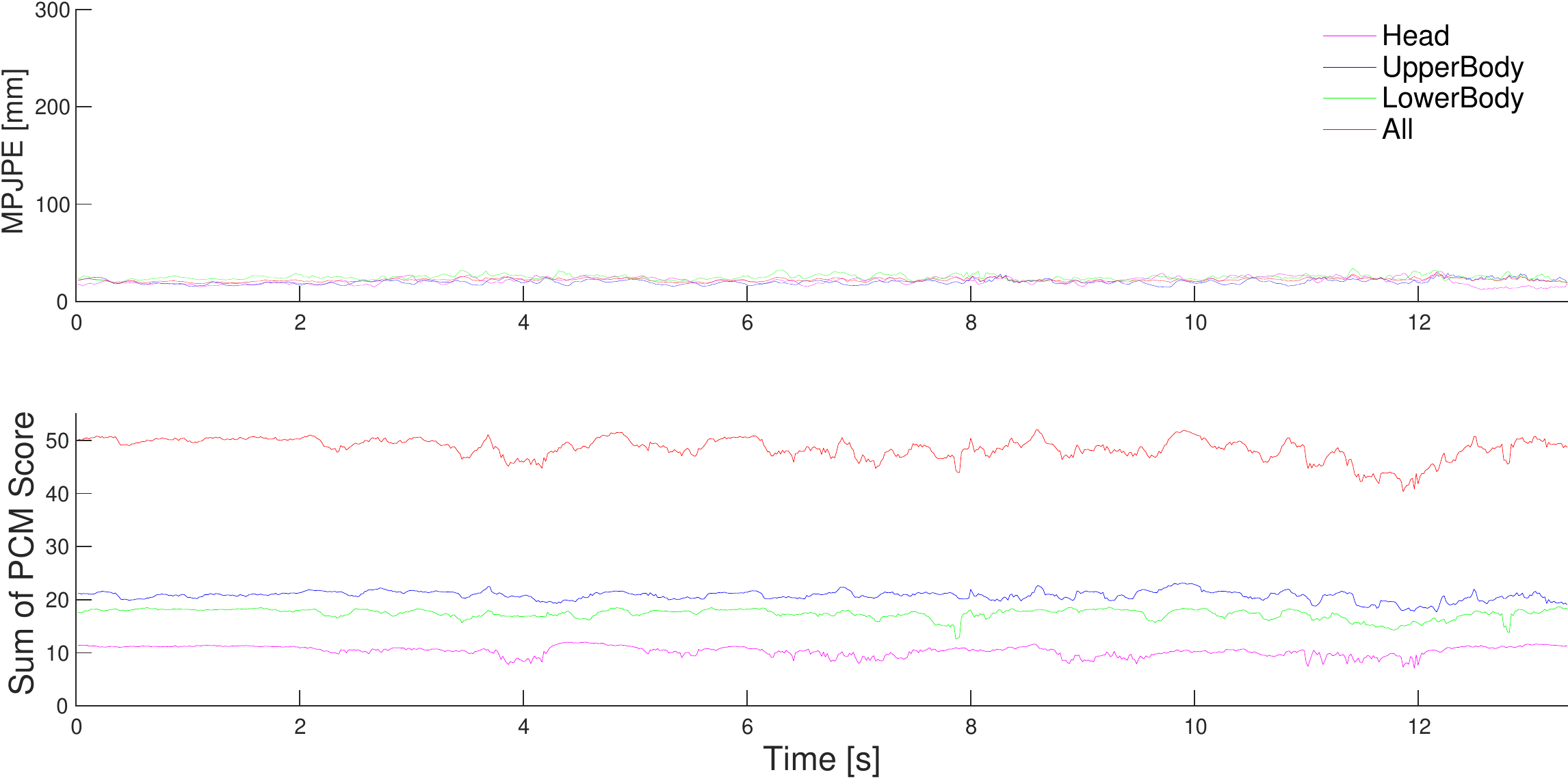}
\caption{MPJPE \& PCM Score, Football Juggling}
\label{fig:test11_Score}
\end{center}
\end{figure}

From the results, in the regular motions such as Dance, Spin Jump, and Matrix, proposed method can estimate joint positions properly and achieves the accuracy of about 30mm in MPJPE.
Fig.\ref{fig:test11_Score} shows MPJPE \& PCM score change in Football Juggling motion which is the smallest in MPJPE.
Note that PCM score is the value of $^{t+1}W^{n}$ defined in Eq.\ref{eq:PCM_SCORE}.
In every frame, proposed method achieves 25mm in MPJPE and PCM score keeps high.

\begin{figure}[ht]
\begin{center}
\includegraphics[width=\hsize,bb=0 0 660 328]{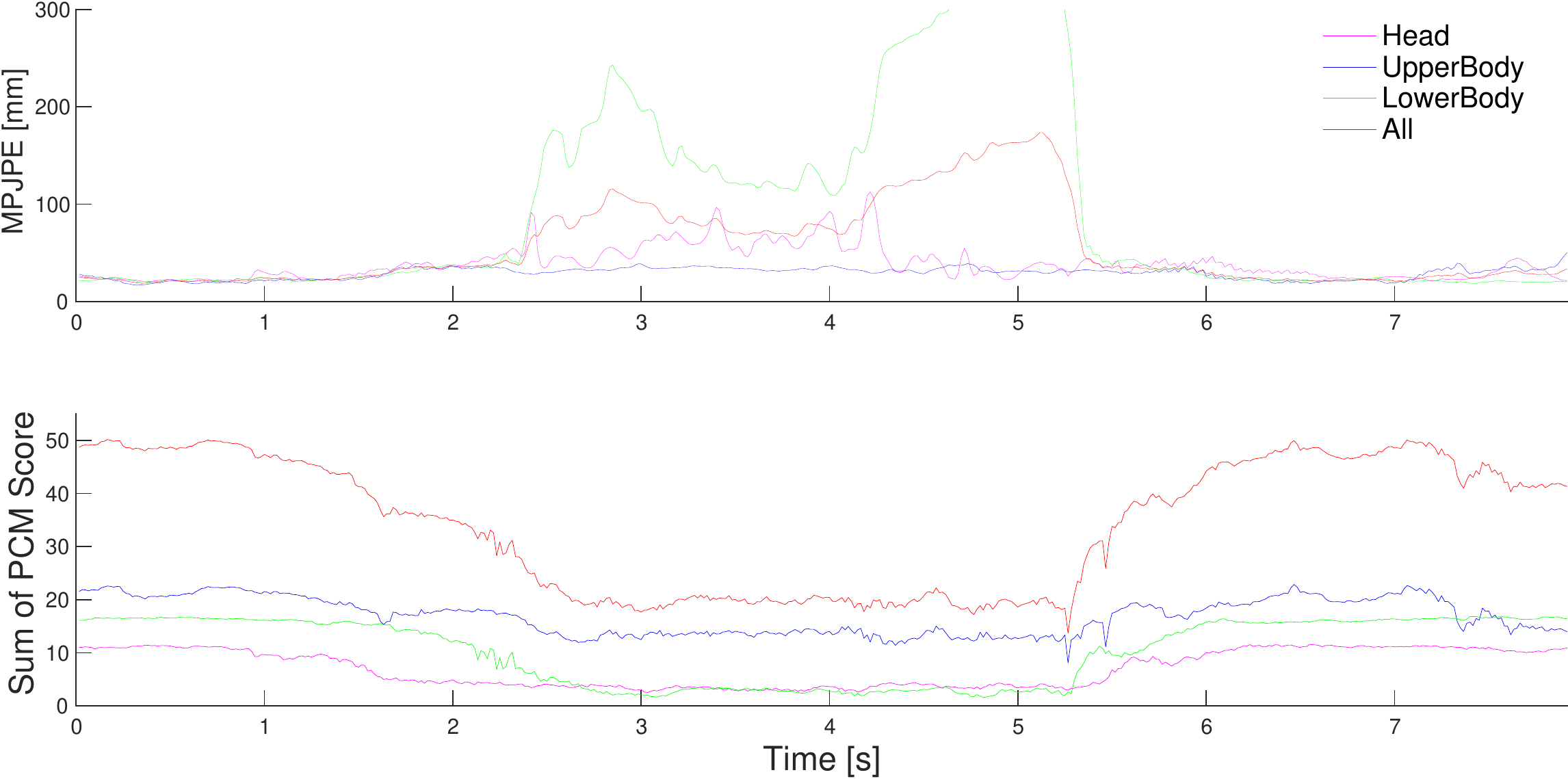}
\caption{MPJPE \& PCM Score, Handstand}
\label{fig:test8_Score}
\end{center}
\end{figure}

However, in inverted motions such as Handstand and Cartwheel, proposed method estimates the joint positions with larger errors, especially in Lowerbody.
Fig.\ref{fig:test8_Score} shows MPJPE \& PCM score change in Handstand motion which is the largest in MPJPE.
As the body tilts, MPJPE became larger and PCM score getting worse.

\begin{figure}[ht]
  \centering
  \includegraphics[width=\hsize]{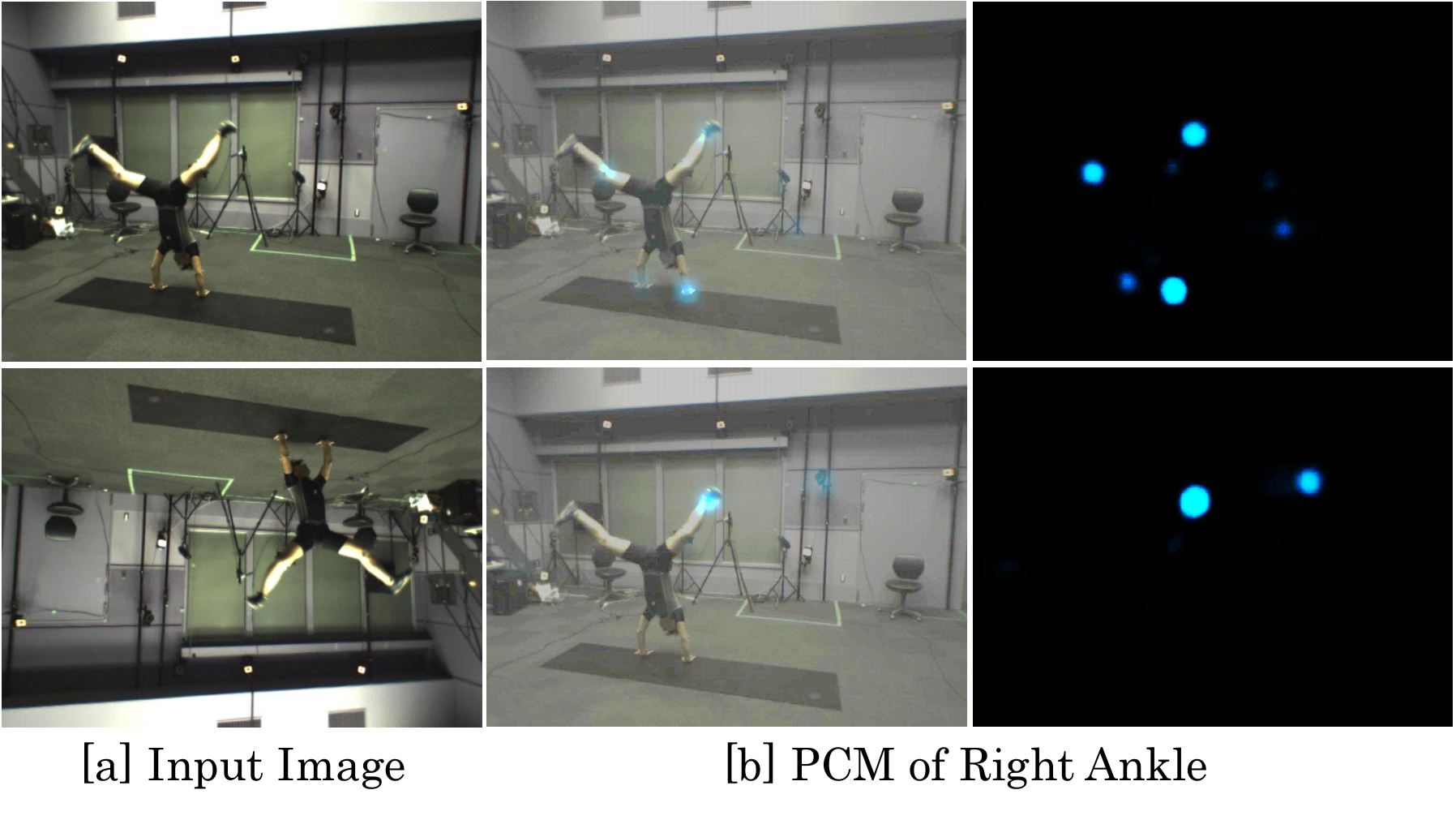}
\caption {PCM Change by Input Image Rotation}
\label{fig:pcm_of_cartwheel_rotate}
\end{figure}

This problem is caused by the accuracy of PCM.
Since OpenPose was created by supervised learning, PCM computation is greatly affected by the training data, and it can be considered that training data did not contain a sufficient amount of inverted posture.
Fig.\ref{fig:pcm_of_cartwheel_rotate} shows the PCM change by the input image rotation.
By inputting rotated image according to the tilt of the human subject, better PCM can be computed.

\subsection{Effect of Input Image Rotation}

\begin{figure}[bh]
 \begin{minipage}[b]{0.99\linewidth}
  \centering
  \includegraphics[width=\hsize]
  {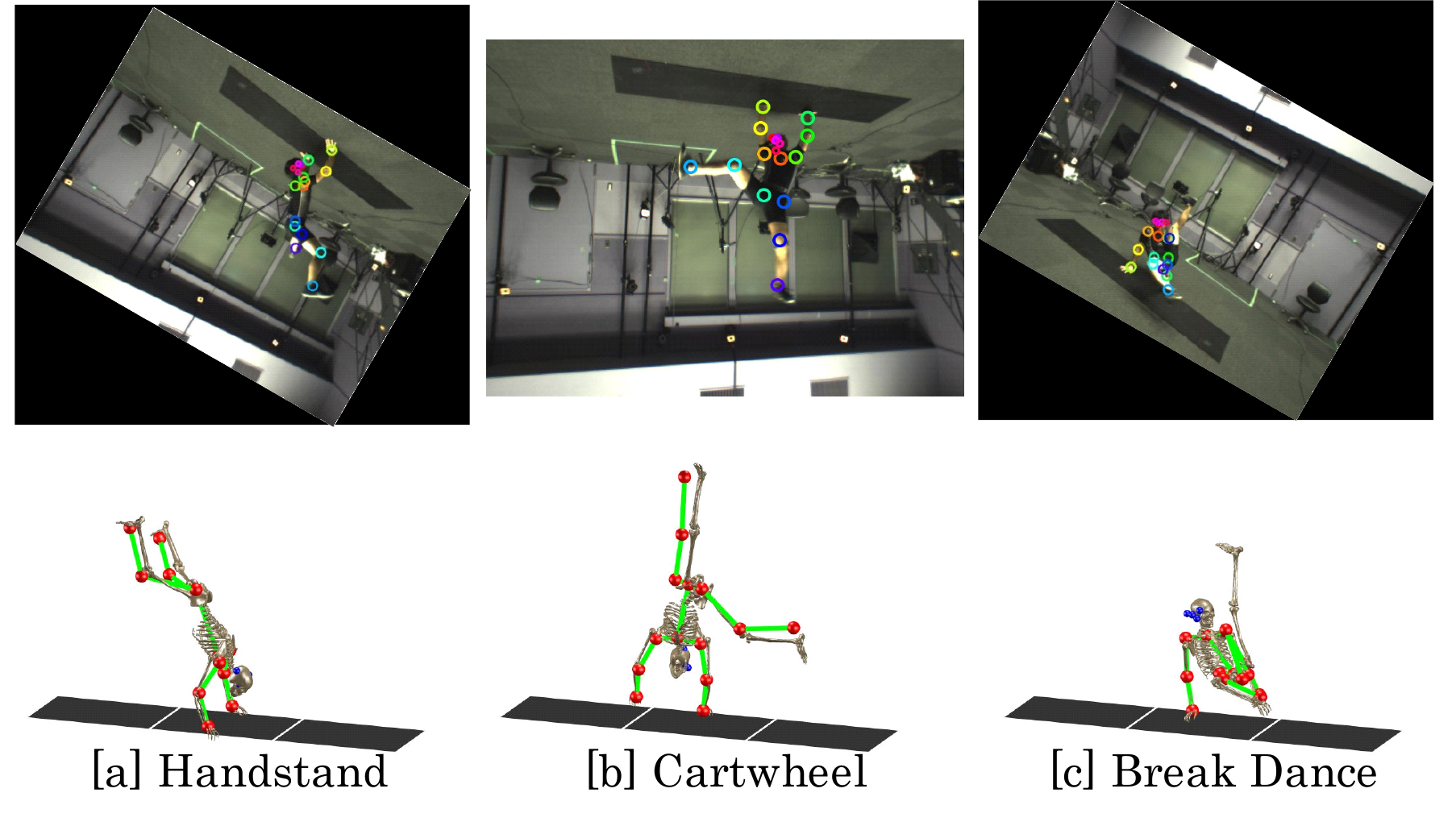}
 \end{minipage}
\caption {Results of Video Mocap with Image Rotation}
\label{fig:output2}
\end{figure}

To improve the MPJPE in inverted motion, we propose computing LowerBody PCM according to the tilt of trunk.
The algorithm is as follows.

From the skeletal model's neck and hip joint position at $t$ frame, the tilt of trunk at $t$ is calculated by each camera image.
According to the tilt, input image at $t+1$ is rotated and then LowerBody's PCM are computed.
PCM of the others are computed from normal input image.
Using these PCM, virtual markers in 3D space are generated and IK computation is performed illustrated as above.

The results of incorporating the input image rotation method into Handstand, Cartwheel, Break Dance are shown as follows.
Fig.\ref{fig:output2} shows that joint position estimation became possible even in inverted motion such as Handstand \& Cartwheel due to the input image rotation.
However, in Break Dance, no significant change appeared.
This can be attributed to the difference between whether the axis of rotation of each motion is perpendicular or parallel to the ground.

\begin{table}[ht]
\centering
\caption{Accuracy Change by Image Rotation}
\label{my-label}
\scalebox{0.98}[1.0]{
\begin{tabular}{|l|l||c|c|c|}
\hline
\multicolumn{2}{|l||}{} & Handstand & Cartwheel & \begin{tabular}[c]{@{}c@{}}Break\\ Dance\end{tabular} \\ \hline \hline
\multirow{2}{*}{\begin{tabular}[c]{@{}l@{}}MPJPE\\ {[}mm{]}\end{tabular}} & LBody & 43.4 (-52.6) & 48.5 (-33.2) & 51.4 (-6.0) \\ \cline{2-5}
 & Total & 36.3 (-19.6) & 40.2 (-12.5) & 39.6 (-2.3) \\ \hline
\hline
\multirow{2}{*}{\begin{tabular}[c]{@{}l@{}}3D-PCK{[}\%{]}\\ @ 50mm\end{tabular}} & LBody & 76.0 (+3.9) & 85.6 (-1.4) & 82.3 (-0.1) \\ \cline{2-5}
 & Total & 82.5 (+1.4) & 82.4 (-0.7) & 86.9 (±0) \\ \hline
\hline
\multirow{2}{*}{\begin{tabular}[c]{@{}l@{}}3D-PCK{[}\%{]}\\ @ 100mm\end{tabular}} & LBody & 92.2 (+10.8) & 93.0 (+2.9) & 93.0 (+0.6) \\ \cline{2-5}
 & Total & 96.8 (+4.4) & 95.3 (+1.1) & 96.7 (+0.2) \\ \hline
\hline
\multirow{2}{*}{\begin{tabular}[c]{@{}l@{}}3D-PCK{[}\%{]}\\ @ 150mm\end{tabular}} & LBody & 95.9 (+9.7) & 96.1 (+4.1) & 95.9 (+0.6) \\ \cline{2-5}
 & Total & 98.4 (+3.6) & 98.0 (+1.7) & 98.2 (+0.3) \\ \hline
\end{tabular}
}
\end{table}

\begin{figure}[H]
\begin{center}
\includegraphics[width=\hsize,bb=0 0 660 328]{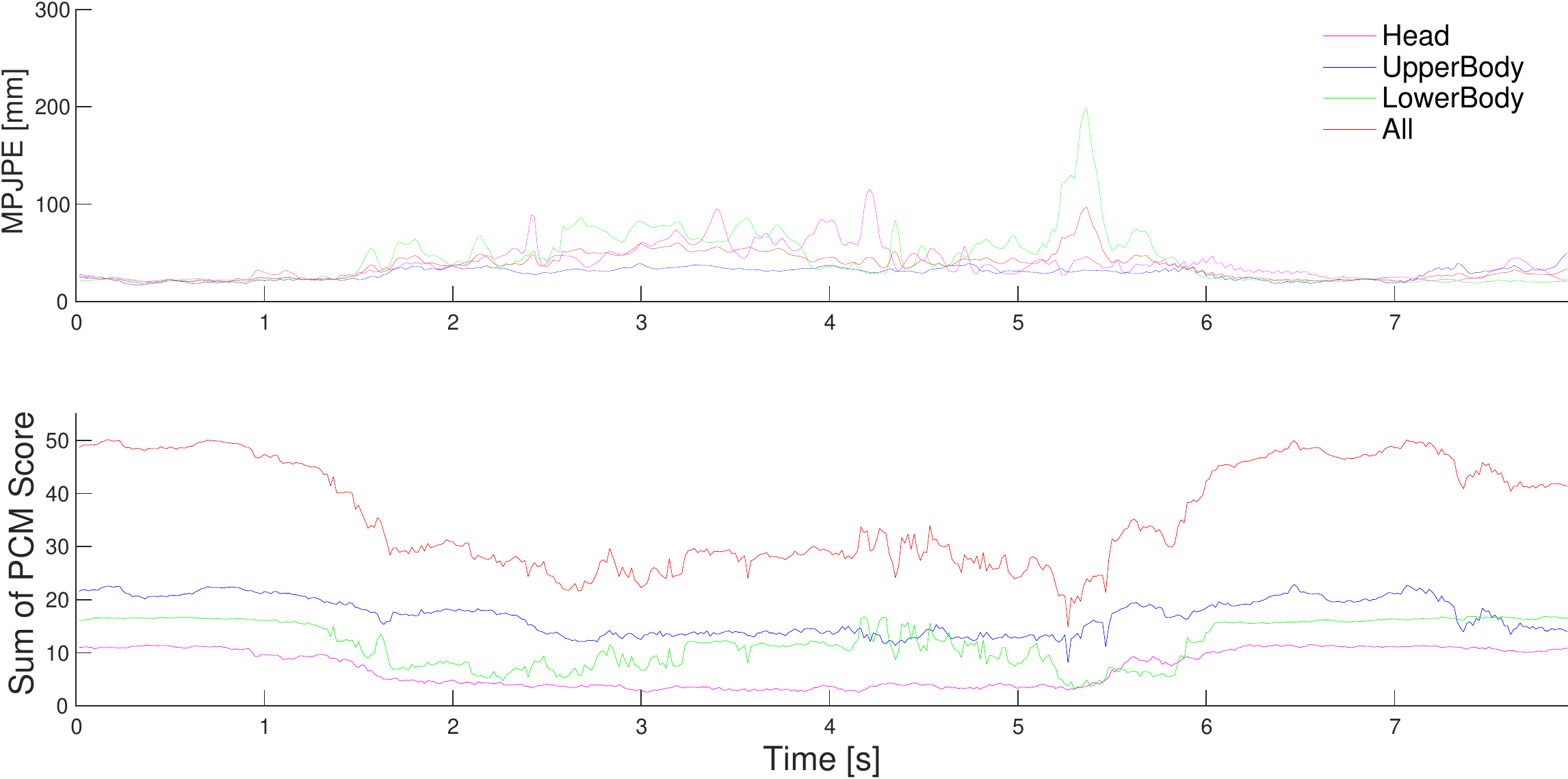}
\caption{MPJPE \& PCM Score: Handstand with Rotation}
\label{fig:test8_rotate_Score}
\end{center}
\end{figure}

\begin{figure}[H]
 \begin{minipage}[b]{\linewidth}
  \centering
  \includegraphics[width=0.92\hsize]
  {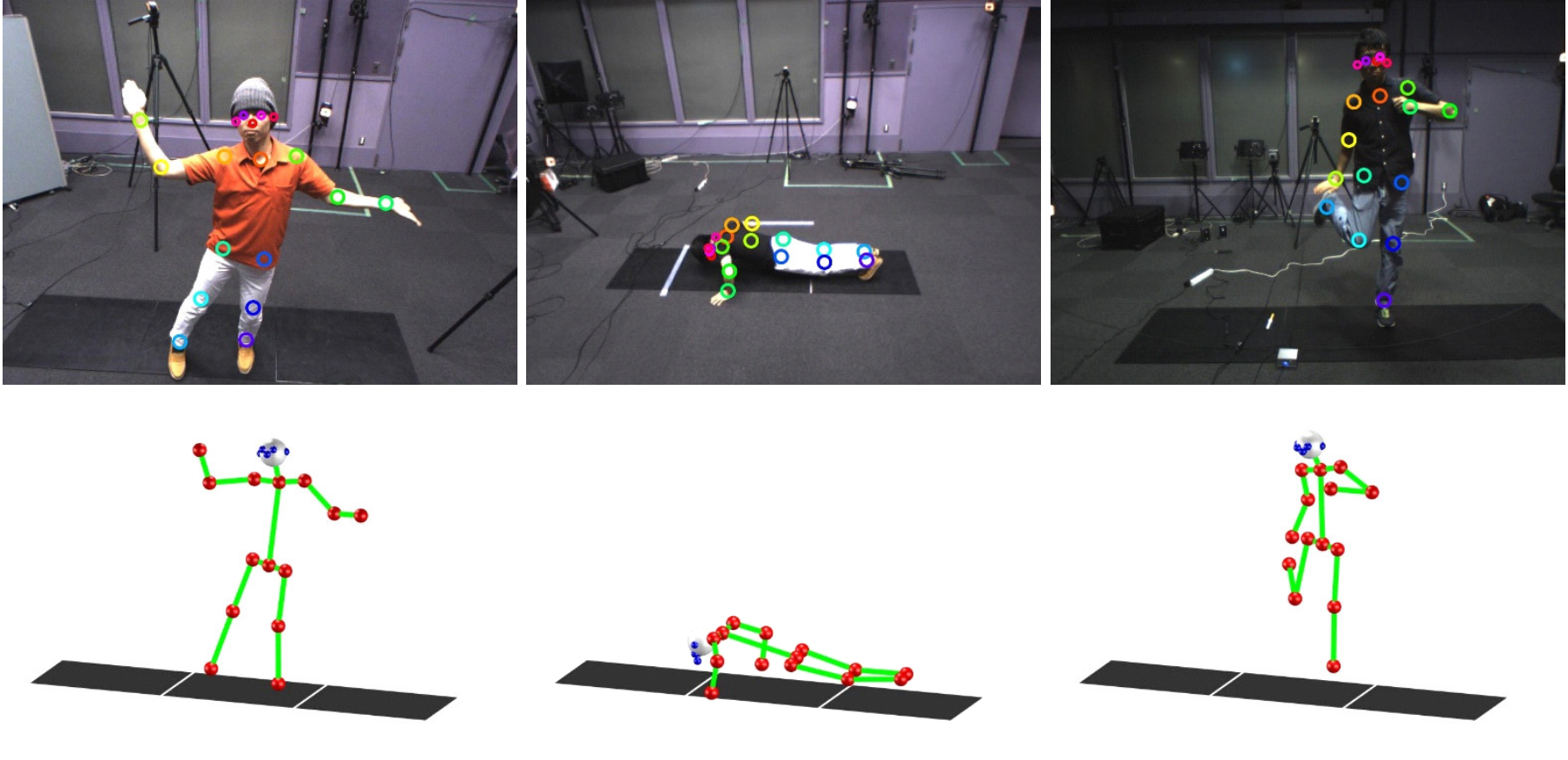}
 \end{minipage}
 \begin{minipage}[b]{\linewidth}
  \centering
  \includegraphics[width=0.92\hsize]
  {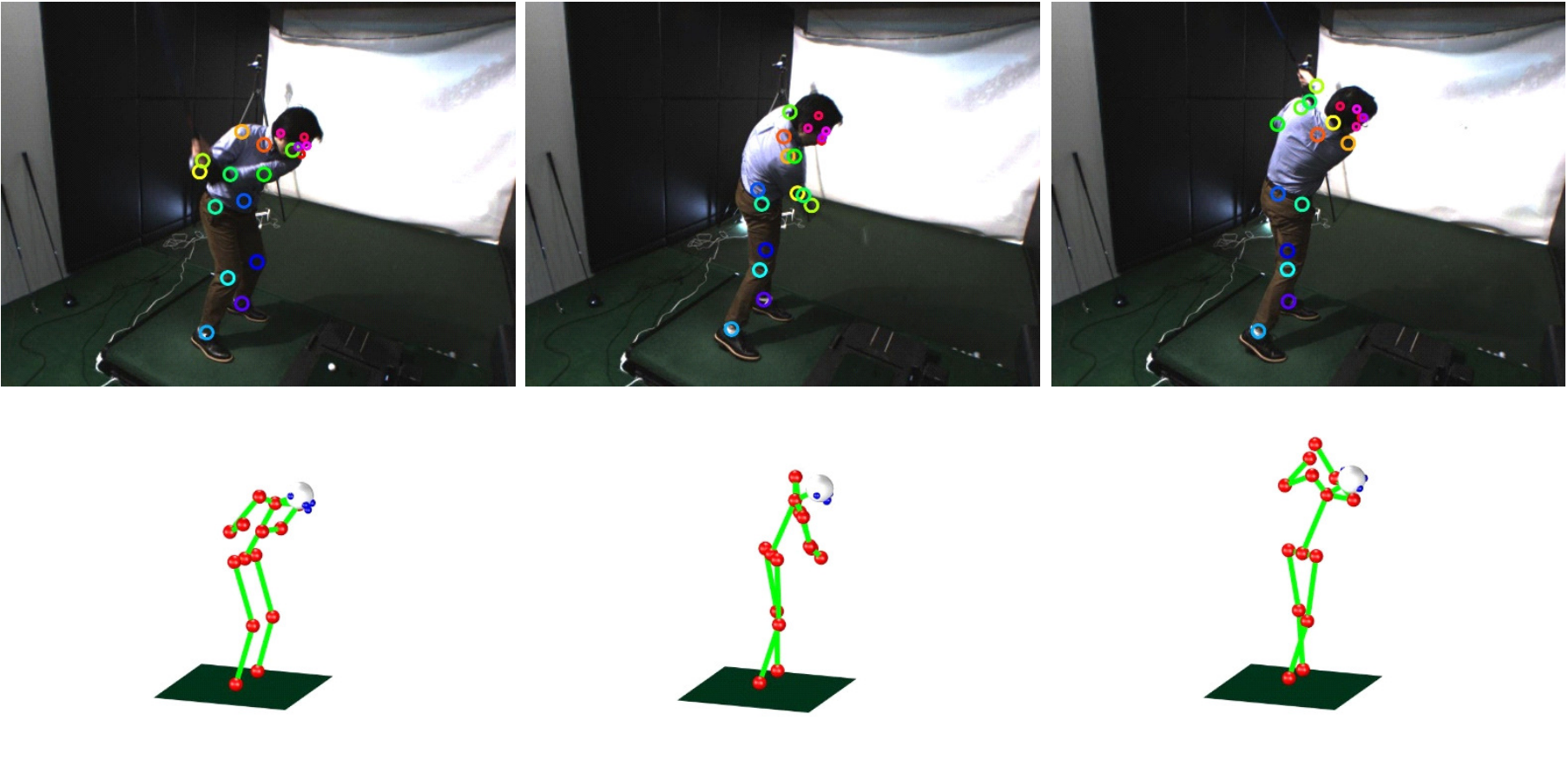}
 \end{minipage}
\caption {Results on Real Environments}
\label{fig:output3}
\end{figure}

To summarize the above results, as shown in the Table.\ref{tab:sum_MPJPE}, we achieved accuracy of 26.1mm of regular motion and 38.8mm of inverted motion in MPJPE. Also, we achieved accuracy of 99.8 \% for regular motions, and 96.4 \% for inverted motions in 3D-PCK at 100mm.
Proposed method is usable regardless of human subject's cloth and environment as Fig.\ref{fig:output3}.

\begin{table}[H]
\centering
\caption{Summary of Accuracy of Video Motion Capture}
\label{tab:sum_MPJPE}
\begin{tabular}{|l|l||c|c|}
\hline
\multicolumn{2}{|l||}{} & Regular Motion & \begin{tabular}[c]{@{}c@{}}Inverted Motion\\ with Rotate\end{tabular} \\ \hline
\multirow{4}{*}{\begin{tabular}[c]{@{}l@{}}MPJPE\\ {[}mm{]}\end{tabular}} & Head & 24.8 & 35.9 \\ \cline{2-4}
 & UpperBody & 29.0 & 31.8 \\ \cline{2-4}
 & LowerBody & 23.4 & 48.3 \\ \cline{2-4}
 & Total & 26.1 & 38.8 \\ \hline
\hline \multirow{4}{*}{\begin{tabular}[c]{@{}l@{}}3D-PCK\\ {[}\%{]}\\ @ 50mm\end{tabular}} & Head & 98.4 & 84.9 \\ \cline{2-4}
 & UpperBody & 93.3 & 87.1 \\ \cline{2-4}
 & LowerBody & 97.5 & 81.2 \\ \cline{2-4}
 & Total & 95.8 & 84.5 \\ \hline
\hline \multirow{4}{*}{\begin{tabular}[c]{@{}l@{}}3D-PCK\\ {[}\%{]}\\ @ 100mm\end{tabular}} & Head & 100 & 97.8 \\ \cline{2-4}
 & UpperBody & 99.7 & 98.9 \\ \cline{2-4}
 & LowerBody & 99.9 & 92.7 \\ \cline{2-4}
 & Total & 99.8 & 96.4 \\ \hline
\hline \multirow{4}{*}{\begin{tabular}[c]{@{}l@{}}3D-PCK\\ {[}\%{]}\\ @ 150mm\end{tabular}} & Head & 100 & 99.3 \\ \cline{2-4}
 & UpperBody & 99.9 & 99.7 \\ \cline{2-4}
 & LowerBody & 99.9 & 96.0 \\ \cline{2-4}
 & Total & 99.9 & 98.2 \\ \hline
\end{tabular}
\end{table}

\section{Conclusion}
The conclusion of this paper is as follows:

\begin{enumerate}
\item An approach of video motion capture was proposed towards image-based 3D human motion reconstruction from multi-camera images using the Part Confidence Maps.
\item A method of spatiotemporal filtering was developed using the human skeletal model for the 3D reconstruction of human motion by mixing temporal smoothing in between two-time inverse kinematic resolutions.
\item The accuracy of the Part Confidence Map was improved using the rotated images for inverted motions such as a handstand and a cartwheel.
\item The experimental results of video motion capture showed that the mean per joint position error was 26.1mm for regular motions and 38.8mm for inverted motions.
When the threshold was set at 100mm as the maximum error of all joint positions for the whole motion, the success rate was 99.8 \% for regular motions and 96.4 \% for inverted motions.
\end{enumerate}

\section{Acknowledgements}

We acknowledge the use of sDIMS, a basic computational library for kinematics and dynamics includes the model of the geometry of human skeletal system, developed in the University of Tokyo.

\begin{footnotesize}
\bibliography{myreference}

\begin{thebibliography}{10}

\bibitem{sdims2}
K.~{Yamane}, Y.~{Fujita}, and Y.~{Nakamura}.
\newblock {Estimation of physically and physiologically valid somatosensory
  information}.
\newblock In {\em IEEE International Conference on Robotics and Automation
  (ICRA)}, 2005.

\bibitem{Murai1}
A.~{Murai}, K.~{Kurosaki}, K.~{Yamane}, and Y.~{Nakamura}.
\newblock {Musculoskeletal-see-through mirror: Computational modeling and
  algorithm for whole-body muscle activity visualization in real time}.
\newblock {\em Progress in Biophysics and Molecular Biology}, 103(2):310--317,
  2010.
\newblock Special Issue on Biomechanical Modelling of Soft Tissue Motion.

\bibitem{Takano:2014}
W.~{Takano} and Y.~{Nakamura}.
\newblock {Synthesis of Whole Body Motion with Pose-Constraints from Stochastic
  Model}.
\newblock In {\em IEEE International Conference on Robotics and Automation
  (ICRA)}, 2014.

\bibitem{MotionA}
{Motion Analysis Corporation}.
\newblock \url{http://www.motionanalysis.com}.

\bibitem{Vicon}
{VICON Corporation}.
\newblock \url{http://www.vicon.com/}.

\bibitem{Shotton1}
J.~Shotton, A.~Fitzgibbon, M.~Cook, T.~Sharp, M.~Finocchio, R.~Moore,
  A.~Kipman, and A.~Blake.
\newblock {Real-time Human Pose Recognition in Parts from Single Depth Images}.
\newblock In {\em IEEE Conference on Computer Vision and Pattern Recognition
  (CVPR)}, 2011.

\bibitem{Tong1}
J.~Tong, J.~Zhou, L.~Liu, Z.~Pan, and H.~Yan.
\newblock {Scanning 3D Full Human Bodies Using Kinects}.
\newblock {\em IEEE Transactions on Visualization and Computer Graphics},
  18(4):643--650, April 2012.

\bibitem{Spinello}
L.~{Spinello}, K.~{O.} Arras, R.~{Triebel}, and R.~{Siegwart}.
\newblock {A Layered Approach to People Detection in 3D Range Data}.
\newblock In {\em Twenty-Fourth AAAI Conference on Artificial Intelligence},
  2010.

\bibitem{Dewan}
A.~{Dewan}, T.~{Caselitz}, G.~D. {Tipaldi}, and W.~{Burgard}.
\newblock {Motion-based detection and tracking in 3D LiDAR scans}.
\newblock In {\em IEEE International Conference on Robotics and Automation
  (ICRA)}, 2016.

\bibitem{Mehta1}
D.~Mehta, H.~Rhodin, D.~Casas, O.~Sotnychenko, W.~Xu, and C.~Theobalt.
\newblock {Monocular 3D Human Pose Estimation Using Transfer Learning and
  Improved CNN Supervision}.
\newblock {\em The Computing Research Repository}, abs/1611.09813, 2016.

\bibitem{VNect:2017}
D.~Mehta, S.~Sridhar, O.~Sotnychenko, H.~Rhodin, M.~Shafiei, H.-P. Seidel,
  W.~Xu, D.~Casas, and C.~Theobalt.
\newblock {VNect: Real-time 3D Human Pose Estimation with a Single RGB Camera}.
\newblock {\em ACM Transactions on Graphics}, 36(4), July 2017.

\bibitem{Kanazawa1}
A.~Kanazawa, M.~J. Black, D.~W. Jacobs, and J.~Malik.
\newblock {End-to-end Recovery of Human Shape and Pose}.
\newblock In {\em IEEE/CVF Conference on Computer Vision and Pattern
  Recognition (CVPR)}, 2017.

\bibitem{Sun1}
X.~Sun, J.~Shang, S.~Liang, and Y.~Wei.
\newblock {Compositional Human Pose Regression}.
\newblock {\em The Computing Research Repository}, abs/1704.00159, 2017.

\bibitem{Wei:2016}
S.-E. Wei, V.~Ramakrishna, T.~Kanade, and Y.~Sheikh.
\newblock {Convolutional pose machines}.
\newblock In {\em IEEE/CVF Conference on Computer Vision and Pattern
  Recognition (CVPR)}, 2016.

\bibitem{Cao:2017}
Z.~Cao, T.~Simon, S.-E. Wei, and Y.~Sheikh.
\newblock {Realtime Multi-Person 2D Pose Estimation using Part Affinity
  Fields}.
\newblock In {\em IEEE/CVF Conference on Computer Vision and Pattern
  Recognition (CVPR)}, 2017.

\bibitem{OpenPose}
{OpenPose}.
\newblock \url{https://github.com/CMU-Perceptual-Computing-Lab/openpose}.

\bibitem{Zhang:2000}
Z.~Zhang.
\newblock {A Flexible New Technique for Camera Calibration}.
\newblock {\em IEEE Transactions on Pattern Analysis and Machine Intelligence},
  22(11):1330--1334, November 2000.

\bibitem{opencv}
Itseez.
\newblock {Open Source Computer Vision Library}.
\newblock \url{https://github.com/itseez/opencv}.

\bibitem{Ayusawa1}
K.~Ayusawa and Y.~Nakamura.
\newblock {Fast inverse kinematics algorithm for large DOF system with
  decomposed gradient computation based on recursive formulation of
  equilibrium}.
\newblock In {\em IEEE/RSJ International Conference on Intelligent Robots and
  Systems (IROS)}, 2012.

\end{thebibliography}
\end{footnotesize}

\end{document}